\documentclass[letterpaper, 10 pt, conference]{ieeeconf}  
\IEEEoverridecommandlockouts   
\usepackage{hyperref}
\usepackage{cite}
\usepackage{graphicx}
\usepackage{booktabs}
\usepackage{subcaption}
\usepackage[font=footnotesize]{caption}
\usepackage{amsmath,amssymb,amsfonts}
\usepackage{algorithm}
\usepackage{algpseudocode}
\usepackage{textcomp}
\usepackage{xcolor}
\usepackage[colorinlistoftodos]{todonotes}

\title{\LARGE \bf
Erwin: Active Robotic Perception for Disease Detection and Mapping in Apple Trees
}

\author{
    Hayden Feddock$^{1}$, Francisco Yandun$^{1}$, Sr\v{d}an G. A\'{c}imovi\'{c}$^{2}$, Abhisesh Silwal$^{1}$
\thanks{$^{1}$ H. Feddock, F. Yandun, and A. Silwal are with the Robotics Institute, Carnegie Mellon University, Pittsburgh, PA 15213, USA {\tt\small \{hfeddock, fyandun, asilwal\}@andrew.cmu.edu}}%
\thanks{$^{2}$S. G. A\'{c}imovi\'{c} is with the Department of Plant Pathology, Virginia Polytechnic Institute and State University, Blacksburg, VA, USA. {\tt\small acimovic@vt.edu}}%
}

\begin{document}

\maketitle
\thispagestyle{empty}
\pagestyle{empty}

\begin{abstract}
Large-scale orchard production requires timely and precise disease monitoring, but
manual scouting is too labor-intensive to scale, leaving outbreaks detected late and
tracked only at coarse, orchard-block resolution. We present the autonomous mobile
active perception system, Erwin, for targeted disease detection and mapping in
dormant apple trees, demonstrated on fire blight---one of the most economically
destructive diseases affecting apple production globally. The system integrates
stereo RGB perception with confidence-aware semantic 3D mapping, fusing semantic
predictions into a volumetric occupancy map that jointly tracks occupancy and
per-voxel semantic confidence to produce actionable spatial maps for growers. To
actively refine observations within complex canopies, we evaluate three viewpoint
planning strategies within a unified perception--action loop: a deterministic
geometric baseline, a volumetric next-best-view (NBV) planner that maximizes
unknown-space reduction, and a semantic NBV planner that prioritizes low-confidence
symptomatic regions. Experiments on five simulated trees across rotational
augmentations and 12 field trees from the Penn State Fruit Research and Extension Center show
reliable symptom localization across diverse canopy geometries: the semantic planner
achieves the best or comparable detection accuracy (mAP/max-F1) at 30 viewpoints in both simulation
and the field, while the volumetric planner achieves the highest ROI coverage in both
settings.
\end{abstract}

\begin{keywords}
Agricultural Automation, Field Robots, Mapping, Methods and Tools for Robot System Design, RGB-D Perception
\end{keywords}

\section{INTRODUCTION}
\label{sec:introduction}

Fruit production in the United States faces mounting pressure from increasing labor shortages, growing global food demand, and tightening economic margins. Orchard crops require extensive monitoring to identify emerging diseases, assess canopy health, and guide management decisions. A substantial portion of disease-related crop loss in pome-fruit orchards is attributable to fire blight---one of the most destructive bacterial diseases affecting apple and pear production globally, with annual losses to U.S. growers estimated to exceed \$100 million \cite{Norelli2003FireBlightManagementTwentyFirst}.

Fire blight is caused by \textit{Erwinia amylovora}, a bacteria which overwinters in bark lesions known as cankers and spreads to infect blossoms and young shoots \cite{van1979FireBlight}. Infected shoots develop a characteristic "shepherd's crook" deformation, while infected woody tissue leads to canker formation and, in severe cases, rootstock blight requiring tree removal. Dormant-season pruning remains the most effective mitigation strategy, but as labor availability declines and operational costs rise, growers often lack the capacity for routine scouting at the frequency and spatial resolution needed for timely intervention. Robotic monitoring systems offer a promising solution by enabling autonomous inspection and construction of actionable disease maps, allowing growers to precisely allocate resources while tracking disease trends with higher temporal and spatial accuracy than manual scouting provides.

\begin{figure}[t]
    \centering
    \includegraphics[width=0.8\linewidth]{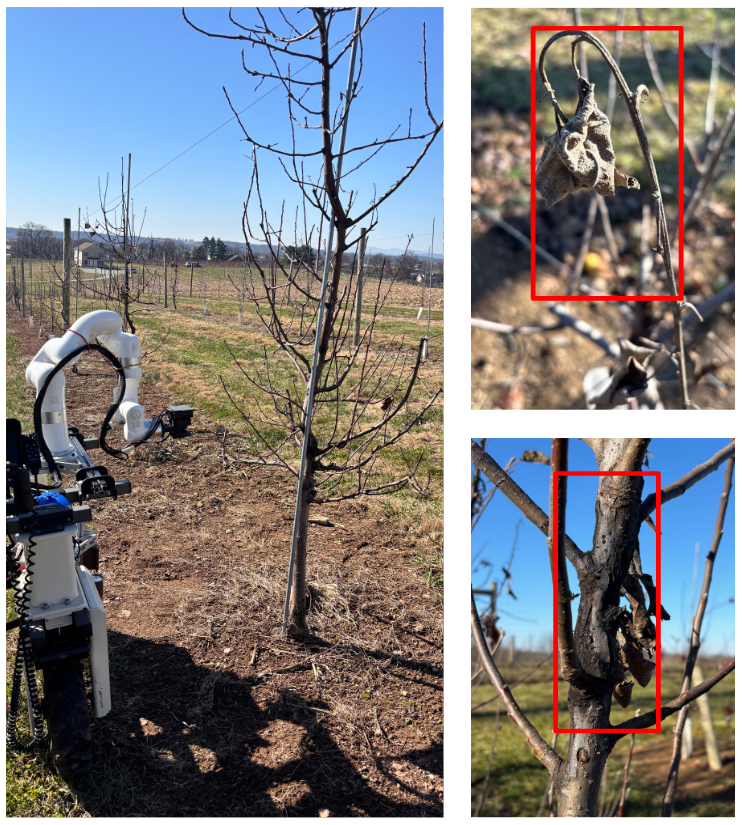}
    \caption{Autonomous disease inspection system (left) and representative fire blight symptoms. Shoot blight instances exhibit a characteristic shepherd's crook (top right), while canker regions appear as sunken or darkened lesions on woody tissue (bottom right).}
    \label{fig:system_and_symptoms}
\end{figure}

Although orchard robotics presents clear benefits, the canopy remains a challenging environment for autonomous perception due to canopy variation, severe occlusions, thin branch geometry, and large variations in natural lighting. Traditional sensor configurations struggle to detect fine visual cues such as canker margins or necrotic shoot tips within these complex tree architectures. These limitations motivate an active perception approach in which a robotic manipulator positions a camera to acquire discriminative, task-relevant views, enabling the use of specialized sensors that are highly effective for disease detection but typically operate with narrow fields of view.

In this work, we present a robotic active perception system for targeted disease detection in dormant apple trees. The robotic arm autonomously selects and executes inspection viewpoints within a closed perception–action loop, while the mobile base remains fixed during inspection. The primary contributions of this work are:
\begin{itemize}
\item \textbf{A confidence-aware semantic occupancy representation} that jointly maintains occupancy probability, semantic labels, and prediction confidence within a unified volumetric map.
\item \textbf{An active viewpoint selection framework, centered on a semantic-aware next-best-view (NBV) planner}, that prioritizes low-confidence symptomatic regions for re-observation, evaluated alongside geometric coverage-based and volumetric NBV baselines to enable refined observation of complex canopy structures.
\item \textbf{Development and field validation of the complete robotic system, Erwin} deployed and validated in a real apple orchard at the Penn State Fruit Research and Extension Center, with performance assessed against splat-based 3D reconstructions.
\end{itemize}

\section{RELATED WORK}

\label{sec:related_works}

Prior work on vision-based detection of fire blight symptoms has largely relied on 
multispectral imagery from aerial systems \cite{schoofs2020UAVFireBlightPearSpectral, 
Xiao2022UAVFireBlightAppleMultispectral} and individual close-up leaf or tissue 
hyperspectral detections \cite{Skoneczny2020DetectionFireBlightAppleHyperspectralLeaf, 
Bagheri2018DetectionFireBlightPearHyperspectral}. However, aerial imagery often misses 
symptomatic indicators beneath the tree crown 
\cite{Bagheri2020AerialDetectionFireBlightPear}. Mahmud et al. proposed a two-modality 
approach combining multispectral UAV imagery for canopy screening with a ground RGB 
instance-segmentation model, confirming that UAV scouting can miss infections on the 
side or bottom of canopies and may lack sufficient resolution to quantify infected areas 
\cite{Mahmud2023MaskRCNNFireBlight}. Veres et al. emphasized ground-level scouting for 
symptom-level localization and deployed a pickup-truck-mounted multi-channel (RGB+NIR) 
imaging rig to collect side-view orchard imagery over the growing season 
\cite{veres2024DetectionFireBlightPearMultiChannel}. More recently, Linker and 
Dafny-Yalin showed that dormant-trunk cankers in pear trees can be detected with 
CNN-based object detection from tripod-acquired imagery and used geo-referenced images 
to generate disease maps; however, this remains a fixed-view acquisition pipeline and 
does not explicitly address occlusion-driven view selection 
\cite{Linker2024MappingFireBlightCankersFRCNN}.

Active perception and next-best-view (NBV) planning provide a principled foundation 
for overcoming occlusions and prioritizing task-relevant observations. In agricultural 
robotics, NBV planning has been successfully applied to targeted plant-part perception: 
Burusa et al. proposed attention-driven and semantics-aware NBV strategies for 
efficient search and detection of task-relevant tomato plant parts in cluttered, 
occluded scenes, validating performance in real greenhouse experiments 
\cite{Burusa2024AttentionDrivenNBVPlantParts, Burusa2024SemanticsAwareNBVPlantParts}. 
In orchards, NBV planning has been applied to apple fruitlet sizing using ROI-targeted 
view sampling and attention-guided information gain, supported by multi-resolution 
volumetric mapping to balance occupancy detail with view evaluation speed 
\cite{Freeman2024AppleFruitletSizing}. These works motivate NBV reasoning in plant 
structures, but focus on reconstruction and sizing rather than disease symptom 
refinement and mapping.

In contrast, our work targets dormant-season fire blight symptoms---where sanitation 
and pruning decisions can be made before in-season symptom escalation---by integrating 
learned symptom segmentation with confidence-aware 3D semantic mapping and 
manipulator-based active viewpoint selection for targeted symptom refinement. We 
additionally leverage flash-illuminated sensing to reduce sensitivity to outdoor 
lighting variability; controlled illumination has been shown to improve image 
consistency and reduce training-data requirements in agricultural vision 
\cite{Silwal2021FlashCamera}.

\section{SYSTEM OVERVIEW}

\label{sec:system_overview}

\begin{figure}[t]
    \centering
    \includegraphics[width=0.8\linewidth, trim={0 400 0 300}, clip]{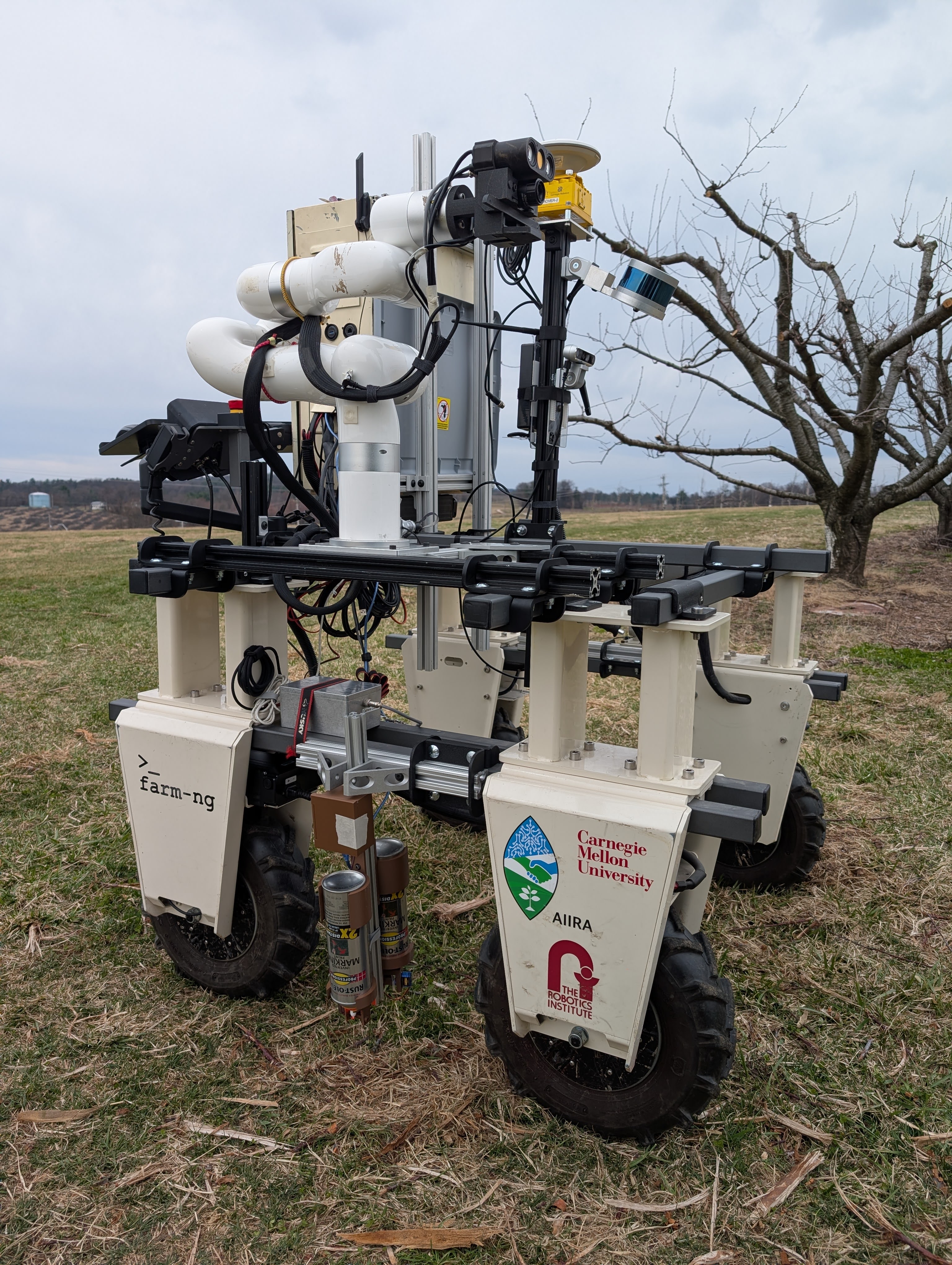}
    \caption{The autonomous disease inspection robot, Erwin, deployed in a dormant apple orchard at the Penn State Fruit Research and Extension Center.}
    \label{fig:field_robot}
\end{figure}

Our platform consists of a mobile manipulation system designed for autonomous disease 
inspection in orchard environments. A 6-DOF xArm6 robotic manipulator is mounted on 
horizontal rails atop an Amiga mobile base, which provides locomotion through orchard 
rows. A custom camera rig, rigidly fixed to the manipulator end effector, houses a 
flash-illuminated stereo RGB camera and an NIR multispectral camera (not used in this 
work). The manipulator enables the camera rig to break the plane of the canopy, 
navigate around occlusions, and capture high-confidence observations of suspected 
disease regions.

Perception, mapping, and viewpoint planning are executed onboard a desktop-grade computing unit 
with an RTX 4070 GPU running ROS2 middleware. The system operates 
in a closed perception--action loop: stereo images are processed to produce semantic 
3D reconstructions, candidate viewpoints are generated based on map state and 
detection uncertainty, and the manipulator is commanded to the next selected 
viewpoint. All sensor data is transformed into a common global reference frame using 
robot odometry and manipulator kinematics.

\section{PERCEPTION AND RECONSTRUCTION}

\label{sec:perception_and_semantic_reconstruction}

\begin{figure*}[t]
    \centering
    \includegraphics[width=0.9\textwidth, trim={0 265 30 0}, clip]{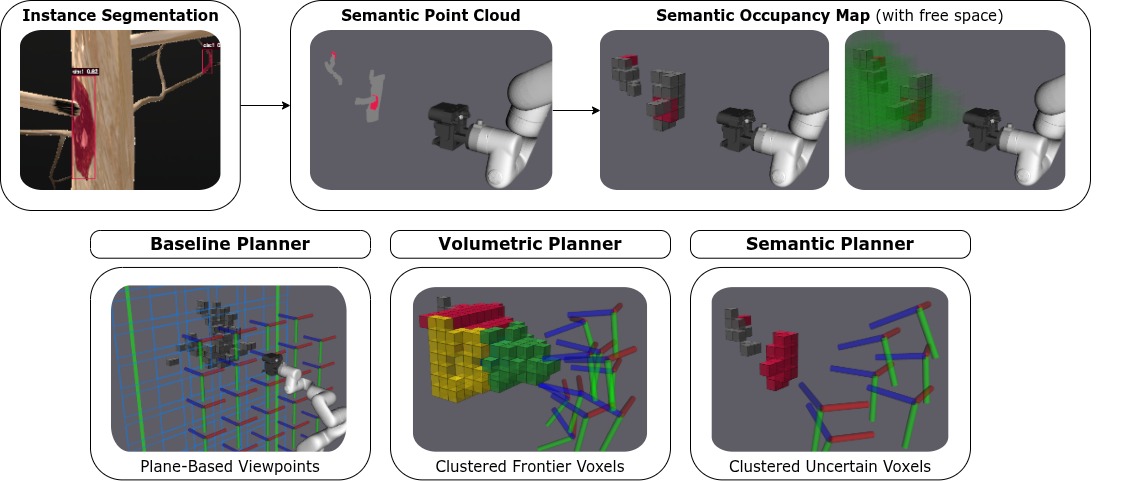}
    \caption{Perception pipeline overview, from flash stereo image acquisition through
    instance segmentation, depth estimation, and semantic point cloud construction to
    integration into a semantic OctoMap.}
    \label{fig:perception_pipeline}
\end{figure*}

The perception pipeline transforms raw stereo RGB observations into a confidence-aware 
semantic 3D representation of disease symptoms. As illustrated in 
Fig.~\ref{fig:perception_pipeline}, we use four stages: (1) flash-illuminated stereo 
image acquisition, (2) instance-level disease detection, (3) real-time depth 
estimation, and (4) semantic point cloud construction. This modular structure allows 
perception components to be independently swapped while maintaining a consistent 3D 
representation for downstream planning.

\subsection{Stereo Flash RGB}
\label{sec:stereo_flash_rgb}

\begin{figure}[t]
    \centering
    \includegraphics[width=0.9\linewidth, trim={0 100 0 200}, clip]{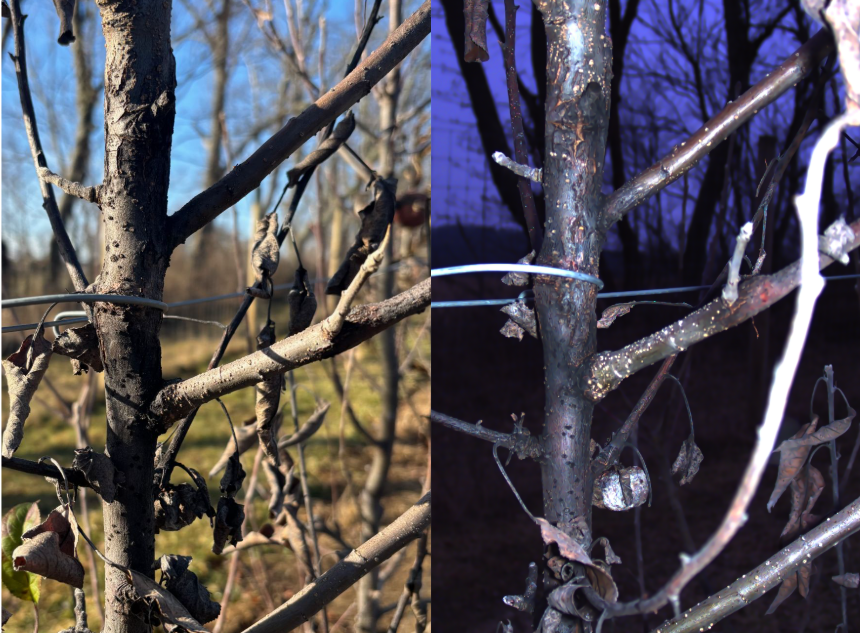}
    \caption{Comparison of a canker imaged in afternoon light with a standard smartphone 
    camera (left) and a flash-illuminated stereo camera (right).}
    \label{fig:flash_contrast}
\end{figure}

Accurate disease detection in orchard environments is hindered by strong illumination 
variability, cast shadows, specular highlights, and dense occlusions within the canopy. 
To mitigate these effects, we employ a stereo RGB camera pair augmented with 
synchronized flash illumination. Flash-illuminated sensing has been shown to reduce 
dependence on natural light sources, mitigate the effects of shadows, and increase 
visual consistency across time of day~\cite{Silwal2021FlashCamera}. We have found that
using flash illumination additionally improves 
contrast along textured surfaces such as canker margins and necrotic shoot tips, as 
seen in Fig.~\ref{fig:flash_contrast}. Images are captured at resolution
$1440 \times 1088$ using a hardware trigger to ensure temporal alignment between left
and right frames.

\subsection{Stereo Matching}
\label{sec:stereo_matching}

Depth estimation is performed using 
FoundationStereo~\cite{wen2025foundationstereozeroshotstereomatching}, a zero-shot 
stereo matching model based on a vision transformer backbone. To meet real-time
requirements, we deploy the small ViT variant with input images scaled to
$448 \times 224$ and compiled via TensorRT, achieving inference times of approximately
30\,ms. The network predicts a disparity map, which is converted to depth using the known stereo baseline and 
focal length. Speckle filtering is applied to remove isolated noise artifacts prior to 
3D projection, producing a dense point cloud in the camera coordinate frame.

\subsection{Detection Model}
\label{sec:detection_model}

Disease symptoms are detected using a convolutional neural network based on the YOLO26
instance segmentation architecture~\cite{yolo26_ultralytics}. The base model was
trained on 1,275 annotated real-world images collected at a commercial orchard distinct
from the test site used for field evaluation, achieving an mAP@0.5 of 0.608 and F1 of
0.61. The model was further fine-tuned on 674 simulation images for simulation-based
evaluation, achieving an mAP@0.5 of 0.760 and F1 of 0.74. The model produces
instance-level segmentations for
visually distinctive dormant-season fire blight indicators, including shepherd's crooks
and canker lesions. For each detected instance, the network outputs a segmentation
mask, class label, and associated confidence score. Predictions below confidence
threshold $\tau$ are discarded.

\subsection{Semantic Point Cloud}
\label{sec:semantic_point_cloud}

To associate semantics with 3D points, instance segmentation masks are rasterized into
per-pixel class and confidence maps, with higher-confidence instances taking priority
over lower-confidence ones in overlapping regions; pixels with no detection are
assigned a background class. Each valid pixel is then projected to 3D and assigned its
class label and confidence from the corresponding map entry. Pixels with invalid
disparity or depth beyond a maximum sensing range are optionally retained as background
points at the maximum range value, allowing the downstream occupancy mapping process to
explicitly represent unobserved space.

\section{SCENE REPRESENTATION}

\label{sec:scene_representation}

Geometric and semantic observations are fused into a derived OctoMap octree at
resolution $r$, where each voxel maintains a probabilistic estimate of occupancy in
log-odds form~\cite{hornung13octomap}, extended with semantic attributes to support
both (i) collision-aware reasoning through occupancy probabilities and (ii)
task-aware reasoning through per-voxel semantic labels and confidences.

\subsection{Occupancy Mapping}
\label{sec:occupancy_mapping}

Given a sensor origin and a point cloud derived from stereo, occupancy measurements
are integrated via ray casting: voxels traversed by rays are updated toward free
space, while ray endpoints are updated toward occupied space. This process yields a
compact volumetric representation that preserves structure in sparse regions while
adaptively refining resolution in cluttered areas.
The map is incrementally updated as new sensor data arrives, restricted to a
$1.0\,\text{m} \times 1.0\,\text{m} \times 2.0\,\text{m}$ bounding region fitted to
the target tree, with ray casting terminated early at the boundary to reduce memory
usage and improve integration speed.

\subsection{Semantic Occupancy Mapping}
\label{sec:semantic_occupancy_mapping}

In addition to occupancy, we store semantic attributes at the voxel level. Each node 
in the semantic octree maintains: (i) the standard occupancy log-odds value, and (ii) 
a semantic class identifier $c \in \mathbb{Z}$ and confidence $s \in [0,1]$.
Semantic labels are integrated into the map using the semantic point cloud, where each 
3D point carries a predicted class and confidence. Semantic updates are applied only 
to voxels that have already been marked as occupied through geometric integration.

Semantic fusion follows the confidence-aware update rule of Burusa et 
al.~\cite{Burusa2024SemanticsAwareNBVPlantParts}, extended here with a per-match 
confidence boost $\gamma$ in addition to the mismatch penalty $\lambda$. Let 
$(c_t, s_t)$ denote the incoming class label and confidence at time $t$, and let 
$(\hat{c}, \hat{s})$ denote the voxel's current semantic estimate. If no prior 
semantic information is stored, we assign $(\hat{c}, \hat{s}) \leftarrow (c_t, s_t)$. 
If the incoming label matches the stored label $(c_t = \hat{c})$, confidence is 
updated as:
\begin{equation}
\hat{s} \leftarrow \mathrm{clip}\!\left(\frac{\hat{s}+s_t}{2}(1 + \gamma),\; 0,\; 1\right),
\end{equation}
reinforcing consistent observations. If the incoming label differs $(c_t \neq \hat{c})$, the label
with higher confidence is retained:
\begin{equation}
(\hat{c}, \hat{s}) \leftarrow
\begin{cases}
(c_t, s_t) & \text{if } s_t > \hat{s}\\
(\hat{c}, \hat{s}) & \text{otherwise},
\end{cases}
\end{equation}
the retained confidence is then penalized to reduce overconfidence under disagreement:
\begin{equation}
\hat{s} \leftarrow \mathrm{clip}\!\left(\hat{s}(1-\lambda),\; 0,\; 1\right).
\end{equation}
Values for $\gamma$ and $\lambda$ are given in Table~\ref{table:params}. This 
fusion strategy favors persistent, high-confidence labels while allowing later 
high-confidence observations to correct earlier errors.

\section{VIEWPOINT PLANNING}

\begin{figure}[t]
    \centering
    \includegraphics[width=0.98\linewidth, trim={380 0 140 265}, clip]{figs/System_Diagram_ral.jpg}
    \caption{Viewpoint candidate generation for the volumetric (left) and semantic
    (right) planners: both cluster voxels of interest---frontier voxels and semantically uncertain
    voxels, respectively---and sample candidate poses on a hemisphere facing
    each cluster centroid.}
    \label{fig:viewpoint_candidates}
\end{figure}

\label{sec:viewpoint_planning}

To evaluate active inspection strategies for unstructured orchard canopies, we implement 
and compare three viewpoint planning approaches: (i) a deterministic geometric baseline 
planner, (ii) a standard volumetric next-best-view (NBV) planner that maximizes occupancy 
uncertainty reduction, and (iii) a semantics-driven NBV planner that prioritizes regions 
of low semantic confidence. All planners operate within the same perception--action loop 
and share a common feasibility filtering stage to ensure kinematic and collision 
validity. Algorithm~\ref{alg:nbv} describes the NBV selection loop shared by the 
volumetric and semantic planners, which differ only in their candidate generator 
$\mathcal{G}(\cdot)$ and information gain function $\mathrm{IG}(\cdot)$. Viewpoint 
candidate generation strategies are illustrated in Fig.~\ref{fig:viewpoint_candidates}.

\subsection{Baseline View Planner}
\label{sec:baseline_view_planner}

The baseline planner provides structured canopy coverage without reasoning about map
uncertainty. A midplane is computed through the defined ROI orthogonal to the primary
viewing direction, positioned at the inspection distance set by the ROI bounding box.
A uniform grid of camera viewpoints is tiled across this midplane, spaced according to
the cross-section of the camera frustum at that distance so that adjacent views overlap
by a specified percentage $\rho$. Viewpoints are executed in a boustrophedon pattern to
minimize manipulator travel. While this approach guarantees deterministic planar
coverage, it does not adapt to tree architectures or utilize map uncertainty.

\subsection{Volumetric-Based NBV Planner}
\label{sec:volumetric_based_nbv_planner}

\begin{algorithm}[t]
\small
\caption{Volumetric Next-Best-View (NBV) Selection}
\label{alg:nbv}
\begin{algorithmic}[1]
\Require Map $\mathcal{M}$, current pose $x_t$, feasible set $\mathcal{Q}$,
candidate generator $\mathcal{G}$, information gain $\mathrm{IG}(\cdot)$, cost $C(\cdot)$
\Ensure Selected next-best view $v_{\text{next}}$
\State Update $\mathcal{M}$ with latest sensor data
\State $V_c \gets \mathcal{G}(\mathcal{M}, x_t)$ 
    \Comment{Generate candidate viewpoints}
\State $V_c \gets V_c \cap \mathcal{Q}$ 
    \Comment{Filter infeasible poses}
\ForAll{$v_i \in V_c$}
    \State $\mathrm{IG}_i \gets \mathrm{IG}(v_i, \mathcal{M})$ 
        \Comment{Information gain from view}
    \State $C_i \gets C(x_t, v_i)$ 
        \Comment{Motion cost}
    \State $U_i \gets \mathrm{IG}_i - \alpha C_i$
        \Comment{Compute utility}
\EndFor
\State $v_{\text{next}} \gets \arg\max_{v_i \in V_c} U_i$ 
    \Comment{Select best viewpoint}
\State Command robot to move to $v_{\text{next}}$
\State \Return $v_{\text{next}}$
\end{algorithmic}
\end{algorithm}

The volumetric planner selects camera poses that maximize reduction of occupancy 
uncertainty within the region of interest, following standard information-gain NBV 
formulations outlined in Algorithm~\ref{alg:nbv}.

\subsubsection{Volumetric Viewpoint Generation}
\label{sec:volumetric_viewpoint_generation}

Candidates are formed from two sources: (i) a midplane grid (as in the baseline 
planner) augmented with small spherical-cap orientation perturbations, and (ii) 
frontier-based viewpoints sampled around clusters of frontier voxels. Here, frontiers 
are defined as free or occupied voxels adjacent to unknown space. To reduce redundancy, 
frontier voxels are clustered using $k$-means (upper limit of 100 voxels per cluster). 
For each cluster centroid, viewpoints are sampled from a hemisphere facing the centroid 
within a bounded radial range near the ideal camera distance.

\subsubsection{Volumetric Information Gain}
\label{sec:volumetric_information_gain}

For a candidate viewpoint $v$, volumetric information gain is computed by ray marching 
through the camera frustum and counting the number of unknown voxels encountered $U_r$, 
averaged over a set of rays $R$:
\begin{equation}
    \mathrm{IG}(v)
    = \frac{1}{|R|}
    \sum_{r \in R} U_r,
\end{equation}
\begin{equation}
    U_r = \sum_{i=0}^{N_r} 
           \mathbf{1}\!\left\{\mathcal{O}(p_{r,i})=\text{unknown}\right\},
\end{equation}
where $p_{r,i}$ is the $i$-th sampled point along ray $r$ and 
$\mathcal{O}(\cdot)\in\{\text{free},\text{occupied},\text{unknown}\}$ is the voxel 
occupancy state. Ray marching terminates at the first occupied voxel, maximum sensor
range, or ROI boundary.

Motion cost $C(x_t, v) = \|x_t - v\|$ is the Euclidean distance from the current
camera pose to the candidate viewpoint, and $\alpha$ is a tunable weight balancing
exploration benefit against motion effort in Algorithm~\ref{alg:nbv}.

\subsection{Semantic-Based NBV Planner}
\label{sec:semantic_based_nbv_planner}

As one of this work's core contributions, the semantic planner extends the
volumetric NBV framework by explicitly targeting regions of low semantic confidence,
prioritizing viewpoints that refine uncertain disease predictions over purely
geometric exploration. The NBV loop, feasibility filtering, and utility formulation
remain unchanged; the key differences lie in candidate generation and the information
gain metric.

\paragraph{Semantic Viewpoint Generation}
\label{sec:semantic_viewpoint_generation}

Candidate generation mirrors the volumetric planner's two sources: (i) the same
midplane grid augmented with small spherical-cap orientation perturbations for broad
canopy coverage, and (ii) hemisphere-sampled viewpoints around clusters of occupied
voxels with low semantic confidence, in place of the geometric frontier clusters used
by the volumetric planner.

\paragraph{Semantic Information Gain}
\label{sec:semantic_information_gain}

The semantic information gain augments volumetric uncertainty with semantic
uncertainty, weighting unknown voxels by $(1-\beta)$ and occupied-voxel uncertainty by
$\beta$:
\begin{equation}
    \mathrm{IG}_{\text{sem}}(v)
    = \frac{1}{|R|}
    \sum_{r \in R} \left( (1-\beta)\, U_r + \beta\, S_r \right),
\end{equation}
\begin{equation}
    S_r = N \sum_{i=0}^{N_r}
           \left(1 - c(p_{r,i})\right)
           \mathbf{1}\!\left\{\mathcal{O}(p_{r,i})=\text{occupied}\right\},
\end{equation}
where $c(p_{r,i}) \in [0,1]$ is the semantic confidence of the occupied voxel at
$p_{r,i}$, $N$ scales $S_r$ onto the same range as $U_r$, and $\beta \in [0,1]$
controls the relative influence of semantic refinement versus volumetric exploration.
Ray marching terminates at the first occupied voxel. Parameter values are reported in
Table~\ref{table:params}.

\section{EVALUATION}

\label{sec:evaluation}

Predicted disease instances are represented as clusters of connected occupied voxels 
belonging to the same semantic class. The centroid of each cluster is computed as the 
mean 3D position of its constituent voxels and used as the predicted symptom location. 
Ground truth symptoms are annotated as 3D segments; the centroid of each annotated 
segment is precomputed and stored as the reference point for matching. One common way to extend standard 2D detection evaluation---which uses bounding-box
IoU as the match criterion---to 3D is to use axis-aligned bounding boxes (AABBs). However, symptom geometry stretches
irregularly along branches without clear start or end points, making AABBs ill-defined.
We instead adopt a centroid-to-centroid formulation, using Euclidean distance between
centroids as the match criterion.

To evaluate detection quality at each step of the planner, predicted clusters and 
ground truth segments are pooled across all trees at each viewpoint index before 
computing the cumulative F1 and mAP, so that symptom-dense trees contribute proportionally 
and per-tree scores aren't undefined on healthy trees with no predictions.

Since voxel cluster boundaries do not always align one-to-one with individual symptoms,
we apply two corrections to predicted clusters prior to matching: clusters spanning
multiple ground-truth points within the evaluation threshold are split into
sub-clusters, correcting for closely spaced symptoms grouped together by voxel
adjacency; and nearby clusters mapping to the same ground-truth point are merged,
correcting for a single symptom fragmented into adjacent detections. Together, 
these corrections ensure that voxelization artifacts at symptom boundaries are not 
mistaken for missed detections or incorrect localizations.

\begin{figure}[t]
    \centering
    \includegraphics[width=0.95\linewidth, trim={400pt 0 400pt 0}, clip]{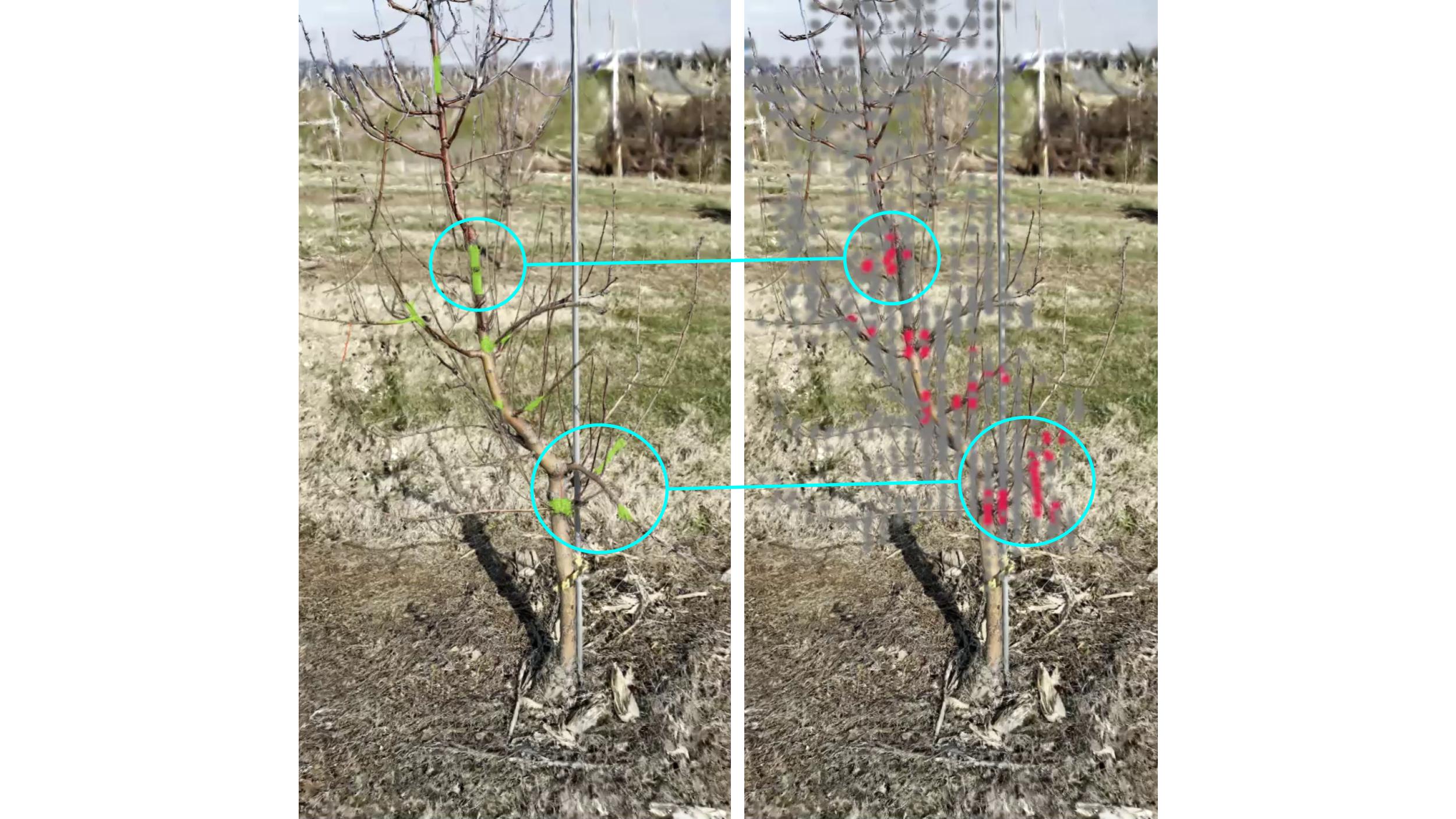}
    \caption{Gaussian splat of a field tree with manually annotated canker symptoms
    (green, left) alongside the canker clusters detected from the OctoMap (red, right),
    with zoomed insets highlighting corresponding regions. The centroid of each
    annotated region serves as the ground truth symptom location for evaluation.}
    \label{fig:splat_annotation}
\end{figure}

\begin{figure}[t]
    \centering
    \includegraphics[width=0.95\linewidth]{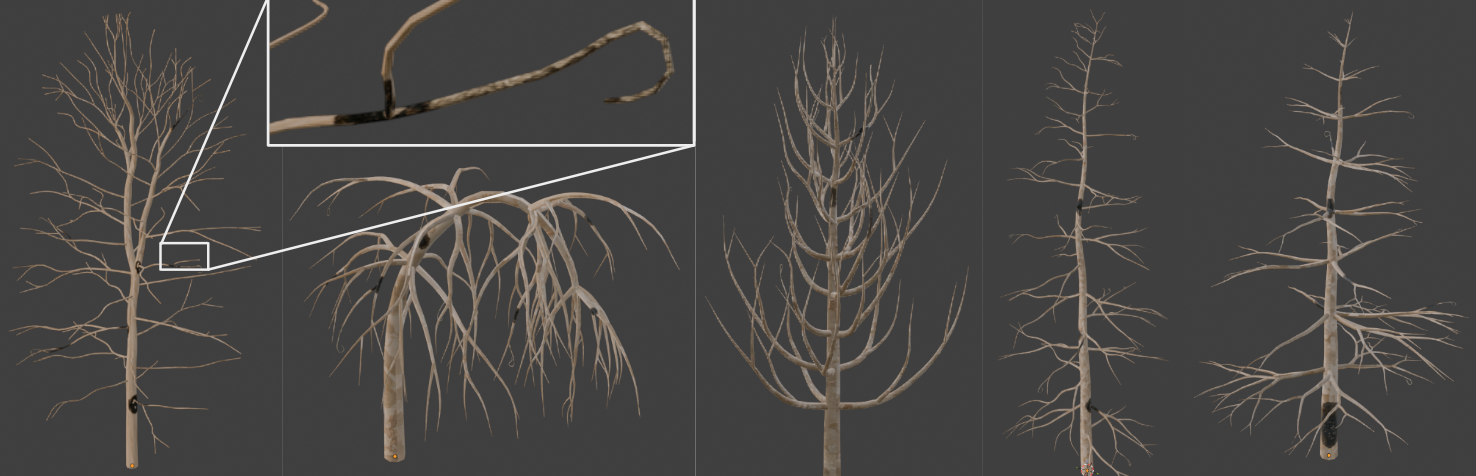}
    \caption{Blender-rendered symptomatic apple trees used in simulated Gazebo 
    evaluations. Each tree contains randomly distributed shepherd's crooks and cankers 
    to simulate realistic outbreak variability and geometric complexity.}
    \label{fig:blender_apple_trees}
\end{figure}

Predictions are swept by descending confidence, matched greedily to ground truth
within distance threshold $t$ (with the splitting and merging described above
applied), and F1 is computed at each threshold; we report the maximum F1 achieved.
The spatial matching threshold is set to $t = 0.10$\,m, selected empirically to
tolerate small centroid localization noise while remaining small enough to
distinguish neighboring symptoms. Max-F1 is evaluated at each viewpoint throughout
planner execution to show how detection quality evolves as new views are acquired.

Mean average precision (mAP) is computed following the COCO detection protocol, with
predictions and ground truth accumulated across all trees using the same matching
procedure as max-F1, applied independently at each distance threshold. The area under
the per-class precision--recall curve gives the per-class AP; mAP averages over 
all classes present in the ground truth or predicted set, and over ten distance 
thresholds from $0.02$\,m to $0.20$\,m in steps of $0.02$\,m. Classes with predictions 
but no ground truth are included in the average rather than excluded, so false 
positives on symptom-free trees are penalized.

To evaluate the effectiveness of each planner at exploring the canopy, we additionally
compute the geometric coverage of the region of interest (ROI), defined as the fraction
of voxels within the bounding box that have been assigned a known (occupied or free)
occupancy state out of the total number of voxels in the ROI. Unlike mAP and max-F1,
which are computed on the pooled cross-tree set, coverage is averaged across trees at
each viewpoint index, since each tree has its own independent OctoMap bounding box.

Ground truth symptom locations were obtained by manually segmenting symptomatic
instances within a shared annotation viewer, with guidance from experts, applied to the Blender mesh geometry for
simulated trees and to a 3D Gaussian splat reconstructed from multi-view imagery for
field trees. In both cases, the centroid of each segmented region was stored as the
ground truth symptom location (Fig.~\ref{fig:splat_annotation}).

Simulation experiments consisted of five symptomatic apple trees generated in Blender 
and evaluated in Gazebo, each containing 5--8 cankers and 5--8 shepherd's crooks 
distributed across varying canopy geometries (Fig.~\ref{fig:blender_apple_trees}).
To increase canopy diversity without requiring
additional tree models, each tree was augmented with rotations of $60^\circ$ increments 
about the vertical axis, yielding up to six distinct orientations per tree and a total 
of up to 30 unique evaluation configurations across the five trees. Field evaluation 
was conducted with the Erwin robot at the Penn State Fruit Research and Extension Center using 12
dormant apple trees, of which 9 exhibited fire blight symptoms and 3 appeared healthy,
providing true negative cases for evaluation. This facility is distinct from the
commercial orchard used to collect training imagery for the detection model, and no
images from the field evaluation site were used during training. In both
environments, trees were positioned $1.0$\,m from the robot base to ensure consistent
sensing conditions. Planning parameters are reported in Table~\ref{table:params}.

\begin{table}[h]
\caption{Planning Parameters}
\label{table:params}
\centering
\small
\setlength{\tabcolsep}{4pt}
\renewcommand{\arraystretch}{0.9}
\begin{tabular}{l c @{\hskip 12pt} l c}
\toprule
Parameter & Value & Parameter & Value \\
\midrule
$r$ (OctoMap resolution) & 0.02\,m & $\lambda$ (mismatch penalty) & 0.2 \\
$\alpha$ (utility weight) & 0.1 & Camera max range & 0.9\,m \\
$\beta$ (semantic weight) & 0.5 & Background confidence & 0.1 \\
$\gamma$ (confidence boost) & 0.2 & Detection confidence & 0.1 \\
\bottomrule
\end{tabular}
\vspace{-2mm}
\end{table}

\section{RESULTS}

\label{sec:results}

Table~\ref{tab:planner_results} summarizes planner performance at viewpoints 15 and
30 across both environments, with per-viewpoint cumulative results shown in
Fig.~\ref{fig:results}.

\begin{table}[h]
\centering
\caption{Planner performance at viewpoints 15 and 30. Full 30-viewpoint inspections
took on the order of 3--5 minutes to complete. Bold marks the top value(s) in each
column.}
\label{tab:planner_results}
\small
\setlength{\tabcolsep}{4pt}
\begin{tabular}{l ccc ccc}
\toprule
 & \multicolumn{3}{c}{Simulation} & \multicolumn{3}{c}{Field} \\
\cmidrule(lr){2-4} \cmidrule(lr){5-7}
Planner & mAP & F1 & Cov. (\%) & mAP & F1 & Cov. (\%) \\
\midrule
\multicolumn{7}{l}{\textit{Viewpoint 15}} \\
Baseline   & 0.167 & 0.273 & 37.88 & 0.089 & 0.124 & 33.77 \\
Volumetric & 0.368 & \textbf{0.504} & \textbf{68.25} & 0.166 & 0.346 & \textbf{56.39} \\
Semantic   & \textbf{0.383} & \textbf{0.493} & 59.00 & \textbf{0.209} & \textbf{0.461} & 48.13 \\
\midrule
\multicolumn{7}{l}{\textit{Viewpoint 30}} \\
Baseline   & 0.366 & 0.465 & 53.75 & \textbf{0.331} & 0.560 & 52.67 \\
Volumetric & 0.429 & 0.537 & \textbf{82.03} & 0.234 & 0.411 & \textbf{71.41} \\
Semantic   & \textbf{0.479} & \textbf{0.588} & 70.21 & \textbf{0.322} & \textbf{0.608} & 60.36 \\
\bottomrule
\end{tabular}
\end{table}

\subsection{Simulation Results}

\begin{figure}[t]
    \centering
    \includegraphics[width=0.98\linewidth]{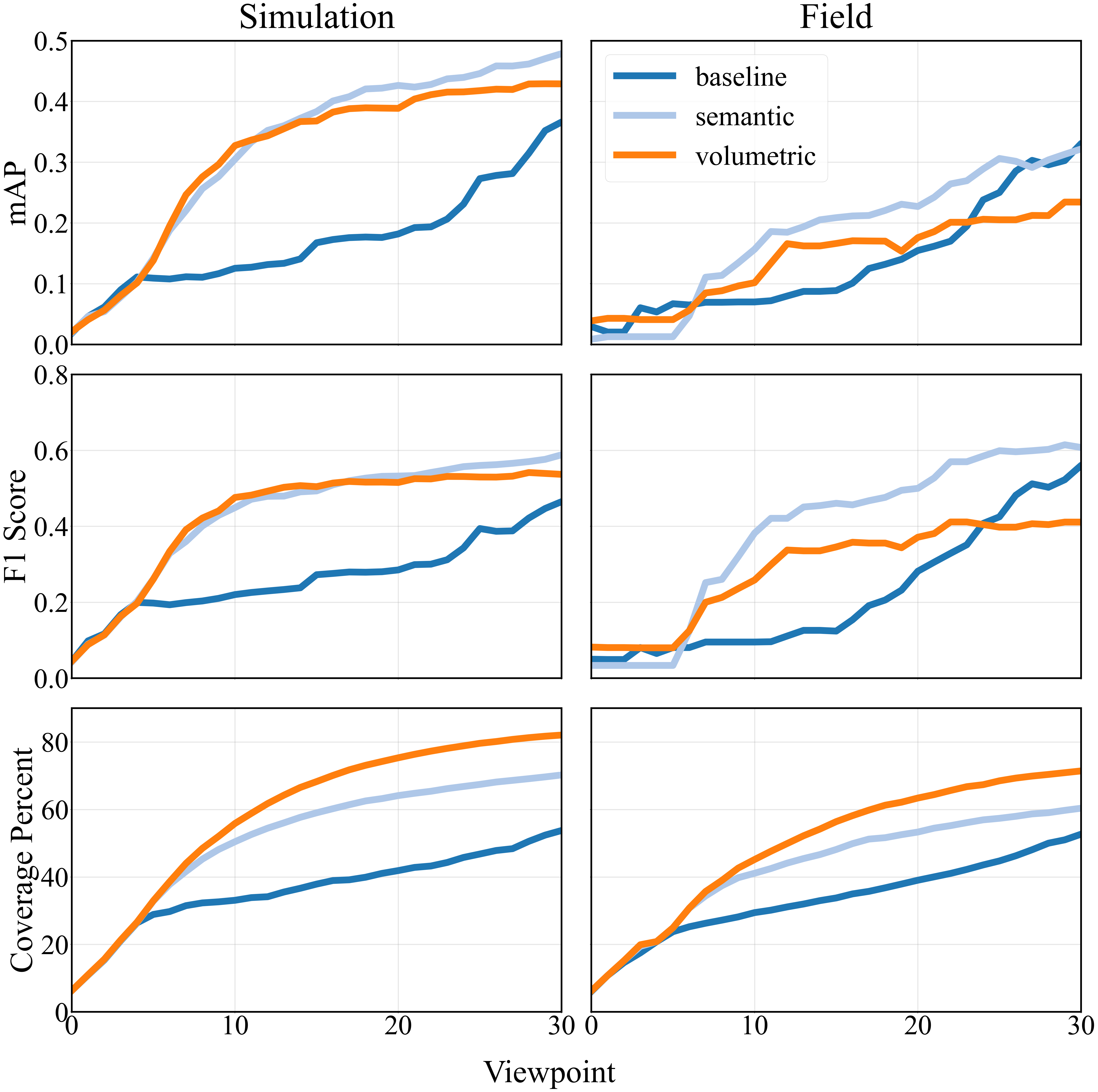}
    \caption{Planner performance averaged across five simulated symptomatic apple trees
    and their $60^\circ$ rotational augmentations (left) and across 12 dormant apple
    trees at the Penn State Fruit Research and Extension Center (right). Curves show mean mAP,
    max-F1, and ROI coverage over 30 viewpoints.}
    \label{fig:results}
\end{figure}

Simulation results are reported across five Blender-rendered trees and their
$60^\circ$ rotational augmentations, yielding up to 30 unique canopy configurations
in total. The volumetric planner achieves the highest ROI coverage ($82.03\%$), while
the semantic planner achieves the highest mAP ($0.479$) and max-F1 ($0.588$).
Fig.~\ref{fig:results} shows the semantic and volumetric planners tracking closely
for most of the sequence, with the semantic planner pulling ahead only in later
viewpoints as it continues refining low-confidence predictions while the volumetric
planner's detection performance plateaus. The baseline planner trails on every
metric, confirming that adaptive viewpoint selection meaningfully improves over
deterministic coverage in unstructured canopies.

\subsection{Field Results}

Field results, collected by deploying Erwin across 12 dormant apple trees, follow the
same pattern: the volumetric planner leads coverage ($71.41\%$) and the semantic
planner leads max-F1 ($0.608$). The semantic and baseline planners reach a
near-identical final mAP ($0.322$ vs.\ $0.331$, within evaluation noise), but at
viewpoint 15 the semantic planner already more than doubles the baseline's mAP
($0.209$ vs.\ $0.089$) and nearly quadruples its max-F1 ($0.461$ vs.\ $0.124$)---
reaching this quality in roughly half the viewpoints, and thus roughly half the
3--5 minute inspection time. This early separation from the baseline is sharper in
the field than in simulation, indicating that targeted refinement of low-confidence
regions is especially valuable under real-world uncertainty.

\section{CONCLUSION AND FUTURE WORK}




\label{sec:conclusions}

We have presented Erwin, an autonomous mobile manipulation system that performs
targeted fire blight detection and confidence-aware semantic mapping in dormant apple
trees, validated in both simulation and a real working orchard despite a domain gap
between training and test orchards.

Across five simulated trees and 12 real trees, adaptive viewpoint selection
meaningfully outperforms deterministic coverage: the volumetric planner provides the
most complete exploration, while the semantic planner matches or exceeds the best
detection accuracy and reaches it markedly faster. Field evaluation was limited to 12
trees, partly because each had to be inspected by all three planners within a single
two-day site visit, and partly because finding fire blight symptoms is inherently
challenging---growers
typically prune out infections as soon as possible, leaving few orchards with the
disease available for study. Together, these results show that Erwin can deliver
fast, fine-grained disease maps under real field conditions, offering growers a
practical path toward timely, targeted intervention at orchard scale.

Future work should focus on autonomous multi-tree traversal to enable full-row orchard
inspection, broader field evaluation across additional orchards and seasons,
integration of learning-based viewpoint selection strategies, and extension to
additional disease cases and multispectral imaging modalities.

\addtolength{\textheight}{-12cm}   





\section*{ACKNOWLEDGMENT}

This work was funded by the USDA National Institute of Food and Agriculture, Specialty Crop Research Initiative project Accession No. 1031507, Grant No. 2023-51181-41319/2023-05675. We would also like to acknowledge Professor Sr\v{d}an G. A\'{c}imovi\'{c} and his team at Virginia Polytechnic Institute and the growers at Rivendale Farms, Trax Farms, and Soergel Orchards for their support.

\label{sec:acknowledgements}



\bibliographystyle{IEEEtran}
\bibliography{references}

\end{document}